\documentclass[letterpaper]{article} % DO NOT CHANGE THIS
\usepackage{aaai2027}
\usepackage[hyphens]{url}  % DO NOT CHANGE THIS
\usepackage{graphicx} % DO NOT CHANGE THIS
\usepackage{natbib}  % DO NOT CHANGE THIS AND DO NOT ADD ANY OPTIONS TO IT
\usepackage{caption} % DO NOT CHANGE THIS AND DO NOT ADD ANY OPTIONS TO IT
\usepackage{algorithm}
\usepackage{algorithmic}
\usepackage{amsmath}
\usepackage{amssymb}

\usepackage{newfloat}
\usepackage{listings}
\DeclareCaptionStyle{ruled}{labelfont=normalfont,labelsep=colon,strut=off} % DO NOT CHANGE THIS
\floatstyle{ruled}
\newfloat{listing}{tb}{lst}{}
\floatname{listing}{Listing}

\usepackage{booktabs}

\title{UniMoCa: Unifying Motion and Camera Controls as Visual Proxies for Faithful Human Video Generation}
\author{
    Liming Tan\equalcontrib\textsuperscript{\rm 1}, Ye Chen\equalcontrib\textsuperscript{\rm 1}, Hao Zhang\thanks{Work done during internship at SJTU.}\textsuperscript{\rm 3}, Lirong Qian\textsuperscript{\rm 1}, Feifei Li\textsuperscript{\rm 2}, Bingbing Ni\corresponding\textsuperscript{\rm 1,2}
}
\affiliations{
\textsuperscript{\rm 1} Shanghai Jiao Tong University
\textsuperscript{\rm 2} USC-SJTU Institute of Cultural and Creative Industry
\textsuperscript{\rm 3} Monash University}

\begin{document}
\nocopyright
\maketitle

\begin{abstract}
Controlling human motion and camera movement is essential for faithful human-oriented video generation, yet remains challenging in multi-person scenes with large body motions, occlusions, and dynamic cameras. Existing pipelines typically rely on visual motion sequences, such as skeleton maps, pose maps, or rendered body representations, for motion control, while using camera embeddings for camera control. Such heterogeneous control interfaces force video generation models to reconcile pixel-aligned visual cues with non-visual geometric embeddings, making motion-camera attribution difficult and sensitive to camera estimation errors. We propose \textbf{UniMoCa}, a representation-driven framework that unifies motion and camera controls in visual space. At the core of UniMoCa is \textbf{Motion-Camera Visual Proxy} (\textbf{MCVP}), a mutually-sharable novel representation that converts 3D human motion and camera trajectories extracted from driving videos into an identity-neutral visual proxy. MCVP renders temporally aligned human geometry under the recovered camera trajectory and augments it with explicit camera trajectory markers, replacing heterogeneous visual-parametric controls with distinguishable visual cues. As both control factors are represented in the same visual space, they become mutually compatible rather than heterogeneous, enabling consistent joint reasoning and editing during video generation. We further curate a \textbf{MCVP-Video} dataset covering complex actions, multi-person interactions, and diverse camera trajectories. Experiments based on the Wan2.2 I2V show that UniMoCa achieves substantial gains in human motion control, camera control, temporal consistency, and camera-aware robustness with minimal additional complexity. More details are shown in our Project page: \url{https://tanliming-daniel.github.io/UniMoCa/}.
\end{abstract}

\section{Introduction}

High-quality video generation is moving beyond visual realism toward precise controllability~\cite{blattmann2023align,blattmann2023stable,ho2022videodiffusion,wan2025wan,yang2025cogvideox,lin2024open, chen2025vectorized, hu2026metaworld, chen2026world}. In human video synthesis, a model should not only preserve natural appearance and temporal coherence, but also follow desired human motions and camera movements~\cite{zhou2025realisdance,he2024cameractrl,cao2025uni3c,zhang2024mimicmotion,wang2025cinemaster}. This requirement becomes particularly challenging in scenes involving multiple people, large body motions, occlusions, and dynamic cameras, where subject motion and viewpoint changes jointly determine the observed video.

\begin{figure}[t]
    \centering
    \includegraphics[width=0.98\columnwidth]{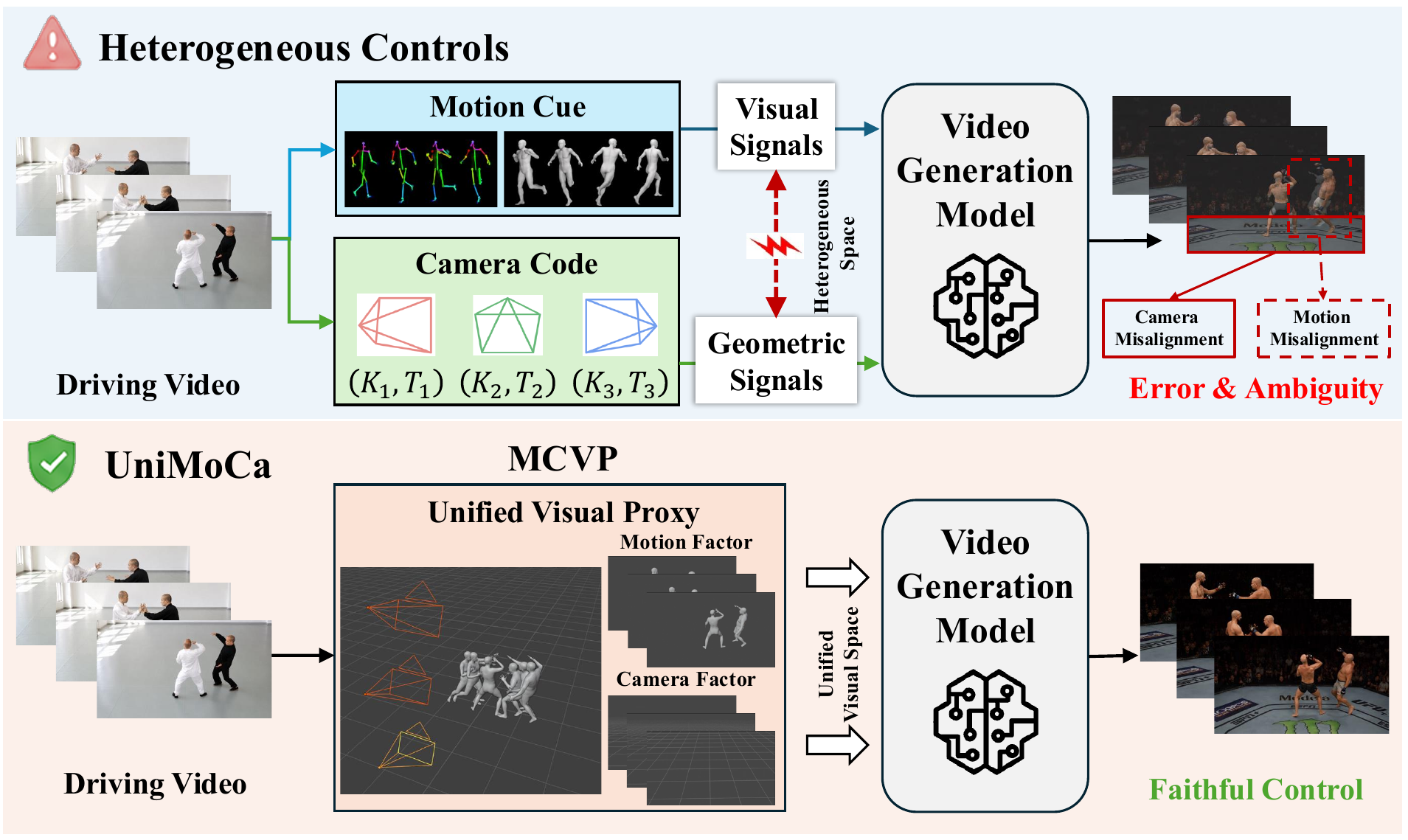}
    \caption{\textbf{Illustration of our Motivation.} Prior methods combine visual motion cues with geometric camera codes, creating ambiguous motion-camera attribution across heterogeneous control spaces (top). UniMoCa instead expresses both controls as MCVP in a shared visual space for faithful motion and camera control (bottom).}
  \label{fig:motivation}
  \vspace{-1.5em}
\end{figure}

Recent controllable human-oriented video generation and character animation methods have made notable progress in motion control ~\cite{zhou2025realisdance,hu2026multianimate,yan2026scail,yan2026scail2,zhou2024realisdance,wang2025unianimate,wan2025wan}. Many approaches introduce visual motion sequences, such as skeleton maps or rendered body representations, to guide human motion, while camera-controllable video generation methods often use Pl\"{u}cker coordinate embeddings or other camera encodings to control viewpoint changes~\cite{he2024cameractrl,bai2025recammaster,cao2025uni3c,luo2025camclonemaster,he2025cameractrl}. These advances improve motion control and camera control \textbf{separately}, but joint camera-and-human motion control remains comparatively underexplored~\cite{cao2025uni3c}. As illustrated in Figure~\ref{fig:motivation}, common pipelines still expose motion and camera through differently structured control signals, making it difficult for the model to learn their respective roles in the generated video.

Specifically, existing motion-camera control pipelines face two related challenges. First, motion control and camera control are often encoded in \textbf{incompatible forms}. Human motion is typically provided as pixel-aligned visual motion sequences~\cite{zhou2025realisdance,hu2026multianimate,yan2026scail,zhou2024realisdance}, whereas camera movement is injected through sparse geometric conditions such as Pl\"{u}cker coordinate embeddings~\cite{he2024cameractrl,cao2025uni3c,li2024controlnet++,zhang2023adding}. Since visual motion sequences are image-space observations already formed under a camera view, they contain camera-projected body changes.The model must therefore align visual motion cues with non-visual camera embeddings and infer whether an observed image change should be attributed to the subject or to the camera, which leads to ambiguity. Second, such \textbf{heterogeneous controls} are sensitive to intermediate estimation errors. Camera trajectory estimation errors propagate into Pl\"{u}cker coordinate embeddings, while visual motion conditions may contain tracking jitter, body reconstruction errors, or rendering misalignment. When these noisy signals are given in different spaces, the model cannot easily determine whether a mismatch comes from camera estimation error or visual motion extraction error. This makes high-quality, temporally stable, and geometrically aligned motion-camera training data a key prerequisite for reliable control.

To address these issues, we propose \textbf{UniMoCa}, a representation-driven framework for faithful human-oriented video generation. Our key idea is not to introduce a more complex control injection module, but to replace the heterogeneous visual-parametric interface with a unified visual representation, accomodating both motion and camera control signal in a consistent visual space. To this end, we introduce \textbf{Motion-Camera Visual Proxy (MCVP)}, a mutually-sharable novel representation. MCVP extracts 3D human motion and camera trajectories from driving videos, renders temporally smoothed and coordinate-aligned human geometry under the recovered camera trajectory, and augments the rendered proxy with explicit camera trajectory markers. In this way, human motion, multi-person spatial relationships, occlusions, and camera dynamics are all exposed to the video generation model as visual cues, while body motion and camera movement remain distinguishable through different visual primitives.

MCVP also mitigates the impact of imperfect intermediate estimation. Unlike prior Pl\"{u}cker-based conditioning~\cite{he2024cameractrl,bahmani2025vd3d,he2025cameractrl}, which injects camera errors as raw perturbations in a parametric space, MCVP converts the recovered camera trajectory into visual proxy frames. As a result, moderate camera errors mainly appear as mild visual deviations in scale, position, viewpoint, or marker location, which video diffusion models can more naturally absorb. Temporal smoothing, coordinate alignment, and sample filtering further suppress high-frequency jitter and failed reconstructions in the recovered human motion and camera trajectories. Since MCVP is identity-neutral, small geometric errors do not directly contaminate the target appearance, but instead act as control noise that the model can learn to handle robustly.

To support this representation, we curate a \textbf{MCVP-Video}  dataset covering complex human actions, multi-person interactions, and diverse camera trajectories. During training, MCVP is encoded into proxy tokens temporally aligned with the target video and jointly processed with video generation tokens by the denoising transformer. We further use shifted positional encoding to distinguish proxy tokens from generated video tokens, allowing the model to treat MCVP as control context rather than generated content. Built on the Wan2.2 I2V~\cite{wan2025wan}, UniMoCa introduces motion-camera control with minimal additional complexity. Experiments show that our method substantially improves human motion control accuracy and camera control accuracy, while also achieving better motion adherence, temporal consistency, and camera-aware robustness in challenging multi-person and moving-camera scenarios.
\begin{figure*}[t]
  \vspace{-2.0em}
  \centering
  \includegraphics[width=\textwidth]{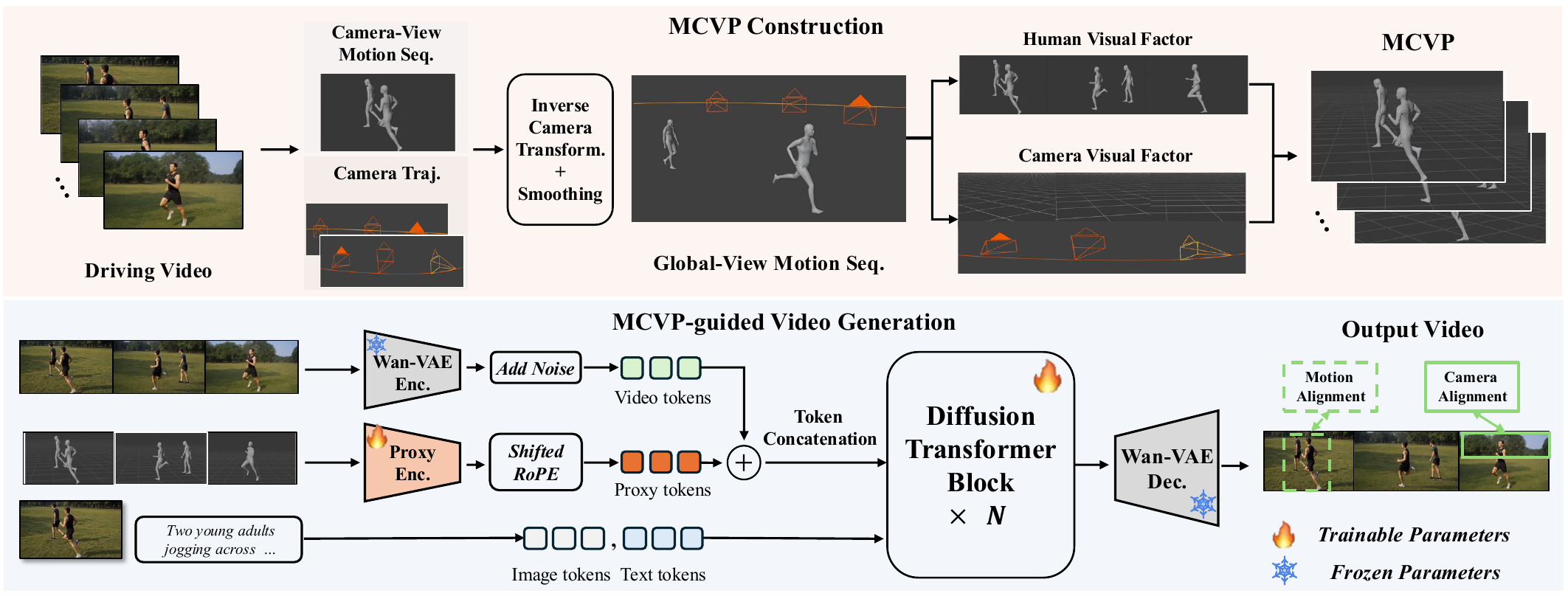}
  \caption{\textbf{Overview of UniMoCa.} Top: we recover camera-view human motion and camera trajectories from a driving video, transform and smooth the motion in global coordinates, and render the human and camera visual factors into MCVP. Bottom: a proxy encoder maps MCVP to visual tokens with Shifted RoPE; these tokens are concatenated with noisy video and appearance/text condition tokens and processed by the diffusion transformer.}
  \label{fig:method-overview}
  \vspace{-1.0em}
\end{figure*}
\section{Related Works}
\noindent\textbf{Human-oriented video generation}
Recent reference-based animation methods condition video diffusion models on human-centric signals to reproduce target performances while preserving subject identity~\cite{hu2024animate,hu2026multianimate,zhang2024mimicmotion,tan2024animate,wang2025unianimate,cheng2025wanani,zhou2024realisdance,zhou2025realisdance, hu2025hunyuancustom}. RealisDance-DiT integrates multiple motion representations~\cite{zhou2025realisdance} including HaMeR~\cite{pavlakos2024reconstructing}, DWPose~\cite{yang2023effective}, and SMPL-X human priors~\cite{pavlakos2019expressive,loper2023smpl,bogo2016keep}, to enhance motion fidelity and geometric consistency. However, it primarily focuses on single-person animation and remains limited in handling multi-character interactions. SCAIL further introduces an identity-agnostic 3D skeleton representation rendered with different hues, enabling scalable multi-character animation without requiring explicit identity annotations~\cite{yan2026scail}. Nevertheless, these methods may suffer from camera-induced jitter, unnatural foot sliding, temporal misalignment, and degraded motion coherence. To address these limitations, SCAIL-2 proposes an end-to-end driving paradigm that directly utilizes driving videos as conditions, avoiding unreliable intermediate representations while preserving rich appearance and motion context. However, its reliance on pre-existing driving videos limits controllability and editability, and the generation quality remains highly dependent on the quality and diversity of the input motion sequences~\cite{yan2026scail2}.

\noindent\textbf{Camera-Controlled Video Generation.}
Camera-controlled video generation aims to manipulate viewpoints while preserving visual consistency and temporal coherence. Existing methods incorporate camera geometry through camera-control branches~\cite{hu2024motionmaster, he2024cameractrl,he2025cameractrl,bahmani2025vd3d,cao2025uni3c} or camera-aware positional encodings~\cite{li2026cameras}, enabling controllable viewpoint synthesis but offering limited control over complex human dynamics. ReCamMaster extends camera control to video re-rendering with novel camera trajectories from a single input video, yet remains dependent on existing videos and provides limited motion controllability~\cite{bai2025recammaster}. Uni3C further introduces joint camera and human motion control using scene point clouds and SMPL-X~\cite{pavlakos2019expressive}, but relies on expensive 3D reconstruction and mainly targets single-person scenarios~\cite{cao2025uni3c}. Achieving efficient camera control with scalable multi-person motion generation therefore remains an open challenge~\cite{zheng2026vidcraft3,zheng2026versecrafter,yu2025trajectorycrafter}.

\section{Method}

Our method is built on a simple observation: precise motion-camera control does not necessarily require a complex control injection architecture; it requires the model to receive motion and camera cues in a form that is both unified and distinguishable. \textbf{UniMoCa} therefore keeps the Wan2.2 I2V~\cite{wan2025wan} largely unchanged and introduces \textbf{Motion-Camera Visual Proxy (MCVP)}, a new visual conditioning representation that converts 3D human motion and camera trajectories into distinguishable visual factors in a dense visual space. Figure~\ref{fig:method-overview} gives an overview of the framework.

\subsection{Preliminaries}

\noindent\textbf{Heterogeneous motion-camera controls.} Given a driving video \(V^d=\{I^d_t\}_{t=1}^{T}\), let \(H_t\) denote the underlying 3D human state at frame \(t\), including body pose, geometry, and multi-person spatial layout, and let \(C_t\) denote the camera state. Existing methods commonly use visual motion control sequences, such as skeleton maps, pose maps, or rendered body representations, to guide human motion. We denote such a visual motion condition by \(\mathcal{V}_t\):

\begin{equation}
\small
\mathcal{V}_t = \mathcal{E}_{\mathrm{mot}}(I^d_t) \approx \psi(\Pi(C_t, H_t)),
\end{equation}
where \(\mathcal{E}_{\mathrm{mot}}\) extracts a visual motion condition from the driving frame, \(\Pi(\cdot)\) denotes camera projection, and \(\psi\) denotes the rendering or rasterization process used to form the visual condition. The important point is that \(\mathcal{V}_t\) is an image- or video-space control signal: it is visually aligned with pixels and already contains camera-projected body observations. In contrast, camera control is commonly represented by Pl\"{u}cker coordinate embeddings:

\begin{equation}
\small
Q_t = \phi_{\mathrm{pl}}(C_t),
\end{equation}
where \(\phi_{\mathrm{pl}}\) maps camera parameters to a sparse geometric embedding. During flow-matching training, the denoising network therefore receives the heterogeneous controls as:

\begin{equation}
\small
u
\left(
z_\tau,\tau,I_0,
\mathcal{V}_{1:T},Q_{1:T},
y;\theta
\right),
\end{equation}
where \(z_\tau\) is the noisy video latent, \(I_0\) is the image-to-video appearance condition, and \(y\) is an optional text condition. Although both \(\mathcal{V}_{1:T}\) and \(Q_{1:T}\) serve as controls, they are produced by heterogeneous mappings: \(\psi\) converts camera-projected human states into pixel-aligned visual maps, whereas \(\phi_{\mathrm{pl}}\) converts camera parameters into sparse geometric ray embeddings. Thus, \(\mathcal{V}_{1:T}\) already reflects camera projection in image space, while \(Q_{1:T}\) remains a non-visual parametric signal. Since supervision is applied only to the target video trajectory, the model must reconcile these heterogeneous cues implicitly, which can weaken motion-camera attribution and make control recombination difficult.

\noindent\textbf{Sensitivity to estimation errors.}
The heterogeneous formulation is also brittle to imperfect intermediate estimates. Let \(\xi^V_t\) denote the extraction or rendering error in the visual motion condition, and let \(\eta^C_t\) denote the camera estimation error. The resulting first-order perturbation to the denoising prediction can be written as:

\begin{equation}
\small
\delta u_\theta
\approx
\sum_{t=1}^{T}
\mathbf{J}^{u}_{\mathcal{V}_t}\xi^V_t
+
\sum_{t=1}^{T}
\mathbf{J}^{u}_{Q_t}\mathbf{J}^{Q}_{C_t}\eta^C_t .
\end{equation}
where \(\mathbf{J}^{u}_{\mathcal{V}_t}\) and \(\mathbf{J}^{u}_{Q_t}\) are the Jacobians of the model prediction with respect to the visual motion condition and the Pl\"{u}cker condition, and \(\mathbf{J}^{Q}_{C_t}\) is the Jacobian of the Pl\"{u}cker embedding with respect to the camera state. The two terms arise from different mechanisms: \(\xi^V_t\) is governed by the upstream visual motion pipeline, such as SMPL estimation, SAM3D-based reconstruction, tracking, and rasterization, whereas \(\eta^C_t\) is transformed by the analytic ray embedding and Pl\"{u}cker coordinate formulation~\cite{yang2026sam,wang2025prompthmr}. Since these errors propagate through different control spaces, the model must implicitly determine whether a mismatch comes from visual motion noise or camera estimation noise. This increases the learning burden of flow-matching training. UniMoCa addresses this issue by replacing heterogeneous visual-parametric inputs with a unified visual representation that makes motion and camera cues distinguishable and converts moderate camera errors into visual-space perturbations.

\subsection{Motion-Camera Visual Proxy}
\label{sec:mcvp}

We propose \textbf{Motion-Camera Visual Proxy (MCVP)}, a visual conditioning representation for motion-camera control. MCVP is constructed from a driving video and organizes motion and camera information into two visual factors: a human visual factor that carries articulated body geometry, and a camera visual factor that carries explicit camera dynamics.

We first extract camera-view human motion from the driving video using SAM 3D Body~\cite{yang2026sam,wang2025prompthmr}:

\begin{equation}
\small
\hat{S}^{\mathrm{cam}}_{1:T} = \mathcal{E}_{\mathrm{body}}(V^d),
\end{equation}
where \(\hat{S}^{\mathrm{cam}}_{1:T}\) denotes a camera-view motion sequence in the 2D+T space. We then estimate camera intrinsics and extrinsics using Depth Anything 3 ~\cite{lin2025depth}:

\begin{equation}
\small
(\hat{K}_{1:T}, \hat{E}_{1:T}) = \mathcal{E}_{\mathrm{cam}}(V^d),
\end{equation}
where \(\hat{K}_t\) and \(\hat{E}_t\) denote the estimated camera intrinsics and extrinsics, respectively. Given the recovered camera parameters, we inverse-project the camera-view motion sequence into a global 3D+T motion sequence and smooth it in the global space:

\begin{equation}
\small
\begin{aligned}
\hat{H}^{\mathrm{glob}}_{1:T}
&=
\mathcal{U}(\hat{S}^{\mathrm{cam}}_{1:T};\hat{K}_{1:T},\hat{E}_{1:T}), \\
\tilde{H}_{1:T}
&=
\mathcal{S}_{\mathrm{czv}}(\hat{H}^{\mathrm{glob}}_{1:T}),
\end{aligned}
\end{equation}
where \(\mathcal{U}\) denotes the inverse camera transformation, and \(\mathcal{S}_{\mathrm{czv}}\) denotes a standard contact-aware zero-velocity trajectory refinement operator~\cite{Zou_2020_WACV}. This step smooths global human trajectories by encouraging contacted foot joints to remain stationary, thereby suppressing temporal jitter and foot-skating artifacts without changing the identity-neutral nature of the proxy.

For notation simplicity, we write \(\hat{C}_t=(\hat{K}_t,\hat{E}_t)\). Instead of injecting \(\hat{C}_{1:T}\) as raw camera parameters, we replay the recovered camera trajectory during rendering. Specifically, we render an identity-neutral, camera-aware human factor \(F^H_t=\mathcal{R}_{\mathrm{body}}(\tilde{H}_t;\hat{C}_t)\), and a non-human camera marker \(F^C_t=\mathcal{R}_{\mathrm{cam}}(\hat{C}_t)\). The final MCVP frame and sequence are:

\begin{equation}
\small
\begin{aligned}
P_t &= \mathcal{G}(F^H_t,F^C_t), \\
P &= \{P_t\}_{t=1}^{T},
\end{aligned}
\end{equation}
where \(\mathcal{G}\) denotes a visual composition operator that overlays the camera marker onto the human visual factor with a fixed spatial layout to form the final proxy frame. Since \(\hat{C}_t\) is replayed during rendering, camera-induced scale changes, viewpoint changes, occlusions, and multi-person spatial relationships are directly visible in the proxy sequence.

MCVP introduces a homogeneous visual interface for motion-camera control. Both \(F^H_t\) and \(F^C_t\) are visual signals: the human factor preserves articulated body geometry, while the camera factor visualizes camera dynamics with explicit markers. Since they are composed into the same proxy frame \(P_t\) and later encoded synchronously with the target video, the denoising transformer observes body motion and camera movement within a shared spatio-temporal visual token space. The visual markers provide an explicit cue for changes in camera parameters, allowing the model to infer camera motion from marker dynamics and implicitly separate camera-induced changes from the body motion cues embedded in \(F^H_t\). In this way, MCVP replaces the heterogeneous interface of visual motion sequences and non-visual camera embeddings with distinguishable visual factors in the same dense visual space.

MCVP also reduces the brittleness caused by imperfect intermediate estimation. Let \(\delta P_t\) denote the visual residual induced by motion and camera estimation. Compared with the heterogeneous perturbation in Eq.~(4), errors in MCVP affect the denoising prediction through a single visual proxy path:

\begin{equation}
\small
\begin{aligned}
\delta u_\theta^{\mathrm{mcvp}}
&\approx
\sum_{t=1}^{T}
\mathbf{J}^{u}_{P_t}
\delta P_t, \\
\delta P_t
&=
\mathbf{J}^{P}_{\tilde{H}_t}\xi^H_t + \mathbf{J}^{P}_{\hat{C}_t}\eta^C_t.
\end{aligned}
\end{equation}

where \(\mathbf{J}^{u}_{P_t}\) is the Jacobian of the model prediction with respect to the visual proxy, while \(\mathbf{J}^{P}_{\tilde{H}_t}\) and \(\mathbf{J}^{P}_{\hat{C}_t}\) describe how errors in the smoothed global motion and recovered camera affect the rendered proxy. The key difference is that both motion and camera errors are absorbed into the same visual residual \(\delta P_t\), so the denoiser only observes a homogeneous image-space perturbation rather than a separate parametric camera shift. In practice, this makes the effect of estimation noise more local and easier to smooth out: errors appear as mild changes in body alignment, scale, viewpoint, or marker location, instead of a global perturbation in a non-visual control branch. Temporal smoothing, coordinate alignment, and sample filtering further suppress high-frequency jitter and failed reconstructions. Since the proxy is identity-neutral, moderate geometry errors do not directly contaminate target appearance, but act as visual conditioning noise that the video model can learn to handle.

\begin{figure*}[t]
  \vspace{-2.0em}
  \centering
  \includegraphics[width=0.95\textwidth]{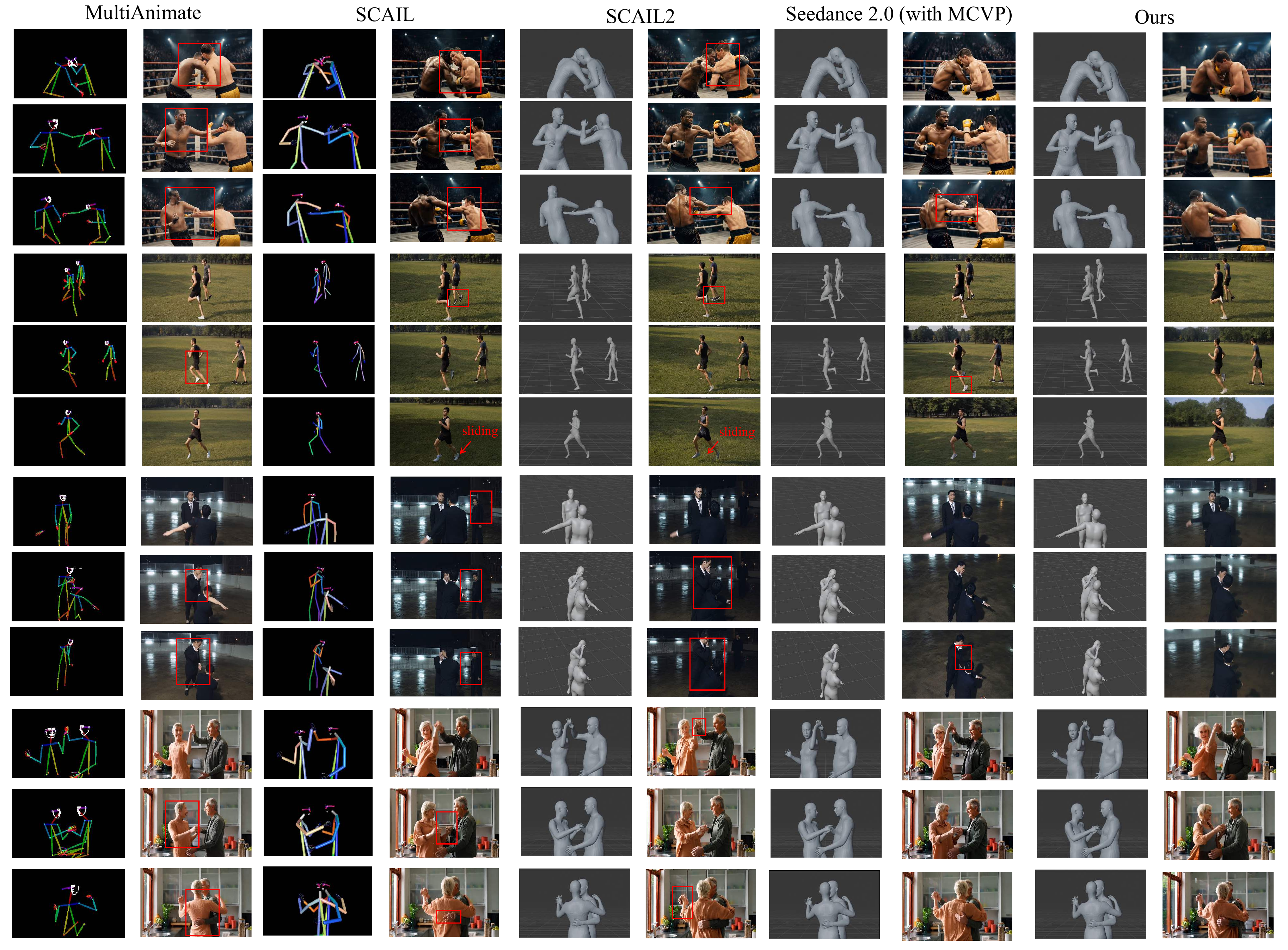}
  \caption{Qualitative comparison of motion control. Note that we also feed MCVP to Seedance 2.0, as it supports this input and performs better than visualized motion alone (see supplementary).}
  \label{fig:motion-comparison}
\end{figure*}

\subsection{MCVP-Guided Video Generation}

\textbf{UniMoCa} uses MCVP as an additional visual input for the Wan2.2 I2V backbone. We do not introduce a complex new control branch or an auxiliary disentanglement loss. Instead, MCVP is encoded as visual proxy tokens and concatenated with noisy video tokens, first-frame tokens, and text tokens in the denoising transformer. Let \(X=\{I_t\}_{t=1}^{T}\) be the training video, \(I_0\) the first-frame appearance input, and \(y\) the text input. The target video is encoded by the Wan VAE and perturbed at flow timestep \(\tau\) to obtain noisy video tokens \(z^{\mathrm{vid}}_{\tau}\), while MCVP is encoded by a trainable proxy encoder as \(z^{\mathrm{mcvp}}=\mathcal{E}_{\mathrm{proxy}}(P)\). The model is optimized with the standard flow-matching objective:

\begin{equation}
\small
\mathcal{L}
=
\mathrm{E}_{x_0,x_1,P,I_0,y,\tau}
\left[
\left\|
u\!\left(
[
z^{\mathrm{vid}}_{\tau};\!
z^{\mathrm{mcvp}};\!
z^{0};\!
z^{y}
],
\tau;\theta
\right)-v_\tau
\right\|_2^2
\right].
\end{equation}
where \(x_0\) and \(x_1\) denote the sampled endpoints of the flow path, \(v_\tau\) is the target velocity at flow timestep \(\tau\), \([\cdot;\cdot]\) denotes token concatenation, and \(z^{0}\) and \(z^{y}\) are the first-frame and text tokens~\cite{lipman2022flow}. The proxy encoder \(\mathcal{E}_{\mathrm{proxy}}\) maps MCVP into the token space of the Wan denoising transformer. In our implementation, we initialize \(\mathcal{E}_{\mathrm{proxy}}\) with the weights of the patchify module in Wan-VAE.

\noindent\textbf{Shifted RoPE for MCVP.} Because MCVP tokens and video tokens share similar spatio-temporal layouts but play different roles, we apply Shifted RoPE to MCVP tokens.~\cite{su2021roformer,heo2024rotary} Let \(t\), \(h\), and \(w\) denote the temporal index, spatial row, and spatial column of a token on the video grid:

\begin{equation}
\small
\begin{aligned}
\mathrm{Pos}^{\mathrm{vid}}_{t,h,w} &= (t,h,w), \\
\mathrm{Pos}^{\mathrm{mcvp}}_{t,h,w} &= (t,h,w+\Delta),
\end{aligned}
\end{equation}
where \(\Delta\) is a fixed spatial offset. Shifted RoPE preserves the temporal correspondence between MCVP and the target video, while marking MCVP tokens as control context rather than generated content. Compared with heterogeneous training, MCVP-guided generation no longer requires the transformer to reconcile a noisy visual motion sequence with a separate non-visual camera embedding. Motion-related and camera-related cues are presented as synchronized visual proxy tokens, allowing the transformer to learn body-motion and camera-motion effects within a unified token space. The separation is induced by the structure of MCVP and the shifted token roles, rather than by an explicit disentanglement objective~\cite{heo2024rotary,chong2025catv2ton,kong2026letrope}.

\begin{table*}[t]
\vspace{-2.0em}
\centering
\small
\setlength{\tabcolsep}{2pt}
\begin{tabular*}{\textwidth}{@{\extracolsep{\fill}}lccccccc@{}}
\toprule
\textbf{Method} &
\multicolumn{4}{c}{\textbf{Automatic Eval. (Video-Bench)}} &
\multicolumn{3}{c}{\textbf{Human Eval.}} \\
\cmidrule(lr){2-5}\cmidrule(lr){6-8}
& \shortstack{\textbf{Img.}\\\textbf{Quality} $\uparrow$}
& \shortstack{\textbf{Motion}\\\textbf{Smooth.} $\uparrow$}
& \shortstack{\textbf{Temporal}\\\textbf{Cons.} $\uparrow$}
& \shortstack{\textbf{Appear.}\\\textbf{Cons.} $\uparrow$}
& \shortstack{\textbf{Img.}\\\textbf{Quality} $\uparrow$}
& \shortstack{\textbf{Motion}\\\textbf{Cons.} $\uparrow$}
& \shortstack{\textbf{Appear.}\\\textbf{Cons.} $\uparrow$} \\
\midrule
MultiAnimate & 3.41 & 3.48 & 3.62 & 3.35 & 3.20 & 3.05 & 2.98 \\
SCAIL1 & 4.01 & 3.86 & 4.08 & 3.95 & 3.74 & 3.62 & 3.57 \\
SCAIL2 & 4.18 & 4.03 & 4.25 & 4.12 & 3.91 & 4.05 & 3.75 \\
Seedance2.0 & \textbf{4.75} & 4.49 & 4.56 & \textbf{4.51} & \textbf{4.31} & 4.17 & \textbf{4.28} \\
\midrule
\textbf{Ours} & 4.26 & \textbf{4.55} & \textbf{4.60} & 4.15 & 4.24 & \textbf{4.21} & 3.91 \\
\bottomrule
\end{tabular*}
\caption{Quantitative comparisons with automatic evaluations and human evaluations.}
\label{tab:motion}
\vspace{-1.0em}
\end{table*}
\begin{figure*}[t]
  \centering
  \includegraphics[width=0.95\textwidth]{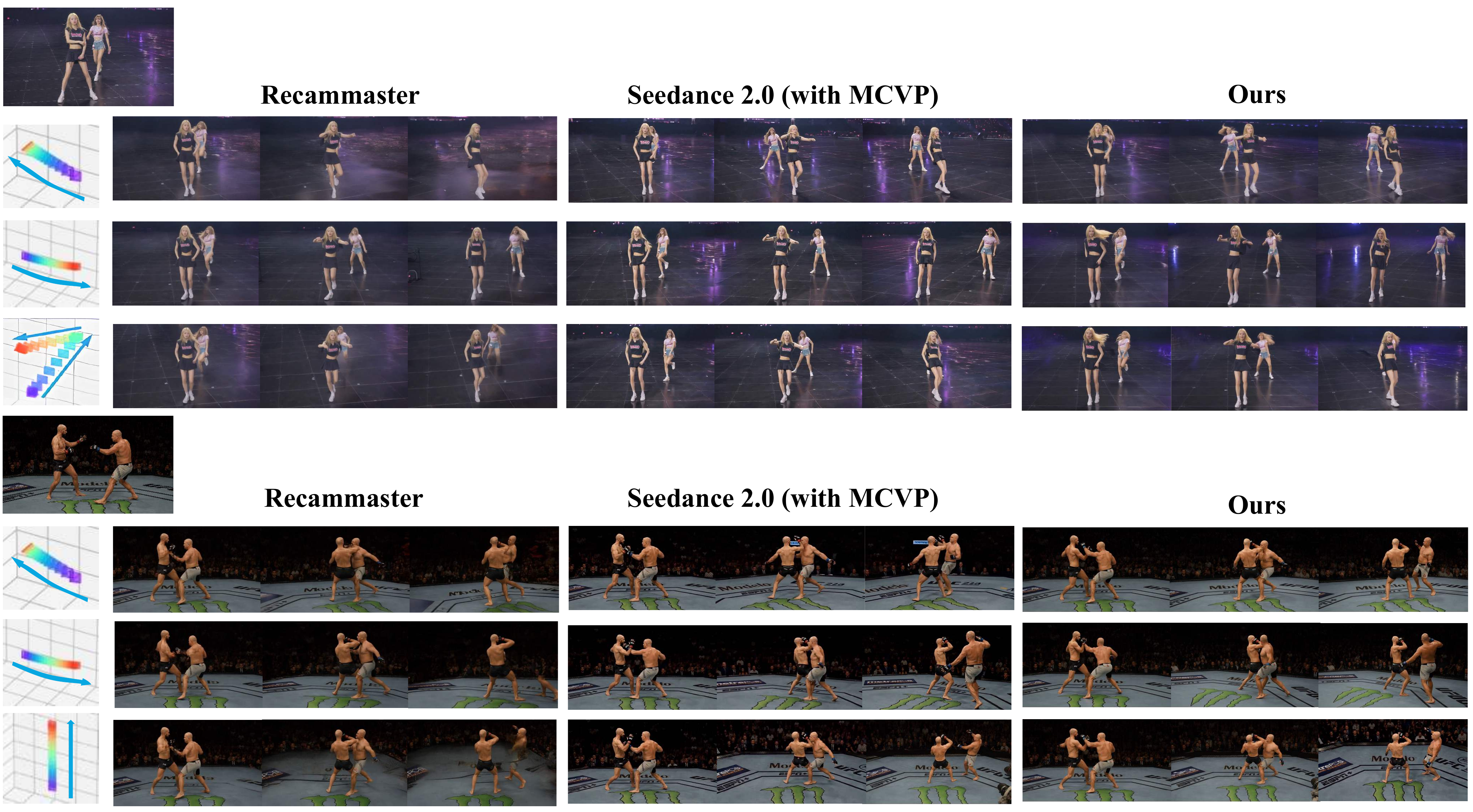}
  \caption{Qualitative comparison of camera control. UniMoCa follows the prescribed reframing more accurately while maintaining subject layout, motion, and scene consistency.}
  \label{fig:camera-comparison}
  \vspace{-1.0em}
\end{figure*}

\noindent\textbf{MCVP Dropout for Robust Motion Learning.} 
While MCVP provides accurate motion and camera guidance, relying on MCVP in every training sample may lead the denoising transformer to overfit the proxy representation and weaken its ability to model intrinsic motion dynamics. To improve robustness, we randomly drop the MCVP input during training with a fixed probability~\cite{srivastava2014dropout}. This stochastic conditioning strategy encourages the model to learn complementary motion priors from the visual context while preserving the controllability provided by MCVP when it is available. This regularization is also intended to reduce sensitivity to incomplete
or imperfect proxy signals; its empirical effect is evaluated in
Table~\ref{tab:ablation}. 

\noindent\textbf{MCVP--Video Data Curation.} 
We collect human-centric videos with continuous shots, prioritizing clips
that contain substantial human motion, multi-person interactions, or dynamic
camera movement~\cite{wang2025koala,wang2024humanvid,zhang2025motion,bai2025recammaster}. Each source video is divided into fixed-length clips, from which we construct temporally aligned MCVP sequences using the procedure
described above. We discard clips with incomplete human reconstruction, discontinuous subject tracks, implausible camera-pose jumps or visible misalignment between the proxy and the source video.
Starting from 200k candidate clips, this pipeline yields 80k valid video--MCVP
pairs, corresponding to approximately 120 hours of training data.

\noindent\textbf{Inference.}
At inference, \textbf{UniMoCa} allows users to customize appearance, human motion,
and camera movement through a common interface. Appearance is specified by
a first-frame image and an optional text prompt, while MCVP can be constructed
from either a driving video or user-authored 3D content. For the latter, users
can author character animations and camera trajectories in Blender or Unreal
Engine, from which we render the human proxy and camera markers to form MCVP.

Because human motion is maintained in a global \(3\mathrm{D}+T\) space, a
motion sequence can be freely combined with different camera trajectories and
target appearances. Users can therefore generate multiple directed shots of
the same performance by changing the camera path, or transfer the same
motion-camera specification across different appearances by changing the first
frame, without model retraining or architectural modifications.

% \subsection{MCVP-Video Dataset Curation}

% Since UniMoCa shifts the main control burden from architectural design to visual representation, the quality of MCVP construction is critical. We therefore build a curated motion-camera training dataset consisting of paired video-MCVP samples. For each training video, we extract 3D human motion, recover the camera trajectory, perform temporal smoothing and coordinate alignment, render the human visual factor under the recovered camera, and compose it with the camera visual factor.

% The dataset construction pipeline filters samples with failed 3D human reconstruction, unstable camera recovery, severe temporal jitter, incomplete human geometry, or unreliable multi-person layouts. This filtering is important because noisy or misaligned proxy-video pairs can teach incorrect correspondence between body motion and camera movement. The final dataset covers complex human actions, multi-person interactions, and diverse camera trajectories, providing the paired supervision required for learning motion-camera control from MCVP.

% With this curated paired dataset, UniMoCa fine-tunes the Wan2.2 image-to-video backbone to receive motion and camera controls through MCVP. The resulting model achieves precise human motion control and camera control with minimal additional complexity, as verified by motion adherence, temporal consistency, and camera-aware robustness in our experiments.

\section{Experiments}
\subsection{Experiment Settings}
\noindent \textbf{Implementation Details.}
We train \textbf{UniMoCa} on our curated MCVP-Video dataset consisting of 80k valid videos for 5000 steps with a batch size of 24 and a learning rate of 1e-5 at the resolution of 832x480, using 8 NVIDIA H100 GPUs for approximately a week. The model is optimized using AdamW. We apply a dropout rate of 0.15 to MCVP tokens and set the Shifted RoPE spatial offset $\Delta=120$.

\noindent \textbf{Evaluation Metrics.}
For evaluation, we construct a benchmark by holding out a subset of videos from our MCVP-Video dataset, ensuring that all evaluation samples are excluded from training. Conventional metrics such as FID and FVD assume the availability of ground-truth videos for comparison~\cite{yu2021frechet,ge2024cdfvd,unterthiner2019fvd}. However, in cross-identity animation, the target appearance is deliberately different from the driving subject, rendering these reference-based metrics unsuitable. To provide more meaningful evaluation, we employ Video-Bench's human-aligned protocol, assessing generations along multiple perceptual axes with scores ranging from 1 to 5~\cite{ning2025video}. Beyond these perceptual measures, we quantify camera control precision using \textbf{RotErr} and \textbf{TransErr}, which measure angular and translational deviations from the target camera path~\cite{bai2025recammaster}.

\subsection{Comparisons with Prior Works}
\noindent\textbf{Motion Control.} We compare our method with state-of-the-art human video generation methods (MultiAnimate~\cite{hu2026multianimate}, SCAIL~\cite{yan2026scail}, SCAIL-2~\cite{yan2026scail2}, and the commercial model Seedance 2.0)in terms of motion-control accuracy. The results are shown in Table~\ref{tab:motion}. UniMoCa outperforms all open-source baselines, especially on motion-centric metrics due to the proposed MCVP, which represents human motion and camera dynamics as temporally aligned signals in a unified visual space. Although Seedance 2.0 remains stronger on appearance-centric measurements such as image quality, UniMoCa surpasses it on motion-control metrics, which is nontrivial for a research system competing with a leading commercial model. We also show some qualitative results in Figure~\ref{fig:motion-comparison}. We can see that UniMoCa better follows articulated motion, contact, and multi-person layouts while remaining visually competitive with Seedance 2.0. These results show that our unified visual Conditioning helps the model distinguish subject motion from viewpoint change while synthesizing both coherently.
\begin{table}[t]
\centering
\small
\begingroup
\setlength{\tabcolsep}{1.5pt}
\renewcommand{\arraystretch}{1.10}

\begin{tabular}{@{}lcc|ccc@{}}
\toprule
\textbf{Method}
& \multicolumn{2}{c|}{\textbf{Camera Accuracy}}
& \multicolumn{3}{c}{\textbf{Video-Bench}} \\
\cmidrule(lr){2-3}
\cmidrule(lr){4-6}

& \textbf{RotErr} $\downarrow$
& \textbf{TransErr} $\downarrow$
& \shortstack{\textbf{Img.}\\\textbf{Quality} $\uparrow$}
& \shortstack{\textbf{Motion}\\\textbf{Cons.} $\uparrow$}
& \shortstack{\textbf{Subject}\\\textbf{Cons.} $\uparrow$} \\
\midrule

ReCamMaster
& 2.10 & 5.35
& 3.88 & 3.42 & 3.86 \\

Seedance2.0
& 1.25 & 4.74
& \textbf{4.73} & 4.04 & \textbf{4.62} \\

\midrule

\textbf{Ours}
& \textbf{1.13} & \textbf{4.53}
& 4.28 & \textbf{4.57} & 4.58 \\

\bottomrule
\end{tabular}
\endgroup

\caption{Quantitative comparison on camera accuracy and Video-Bench metrics.}
\label{tab:camera}
\vspace{-1.0em}
\end{table}
\begin{table}[t]
\centering
\small
\begingroup
\setlength{\tabcolsep}{4pt}
\renewcommand{\arraystretch}{1.12}
\begin{tabular}{@{}lcccc@{}}
\toprule
\textbf{Variant}
& \textbf{RotErr} $\downarrow$
& \textbf{TransErr} $\downarrow$
& \shortstack{\textbf{Motion}\\\textbf{Cons.} $\uparrow$}
& \shortstack{\textbf{Image}\\\textbf{Quality} $\uparrow$} \\
\midrule
Baseline & - & - & 4.18 & 4.27 \\
+ Pl\"ucker Enc. & 2.86 & 5.33 & 3.95 & 4.08 \\
+ ProPE & 1.94 & 5.14 & 4.08 & 4.16 \\
+ MCVP & \textbf{1.13} & \textbf{4.53} & \textbf{4.21} & \textbf{4.28} \\
\midrule
w/o shifted RoPE & 2.15 & 5.27 & 3.93 & 4.01 \\
w/o Dropout & 1.25 & 4.83 & 4.10 & 4.18 \\
\bottomrule
\end{tabular}
\endgroup
\caption{Ablation study of key components.}
\label{tab:ablation}
\end{table}

\noindent\textbf{Camera Control.} We compare our method with SOTA camera-control baseline ReCamMaster ~\cite{bai2025recammaster} and Seedance 2.0 in terms of camera-control accuracy. The results are shown in Table~\ref{tab:camera}. UniMoCa clearly outperforms the open-source baseline, especially on camera-tracking metrics and motion consistency, thanks to the proposed MCVP. It is noted that UniMoCa surpasses Seedance 2.0 on camera accuracy metrics. We also show some qualitative results in Figure~\ref{fig:camera-comparison}. We can see that UniMoCa better follows the prescribed reframing while preserving subject layout and motion coherence. These results show that unified visual conditioning separates subject motion from camera motion, enabling more accurate camera following while preserving human dynamics.

\begin{figure}[t]
  \vspace{-0.5em}
  \centering
  \includegraphics[width=\columnwidth]{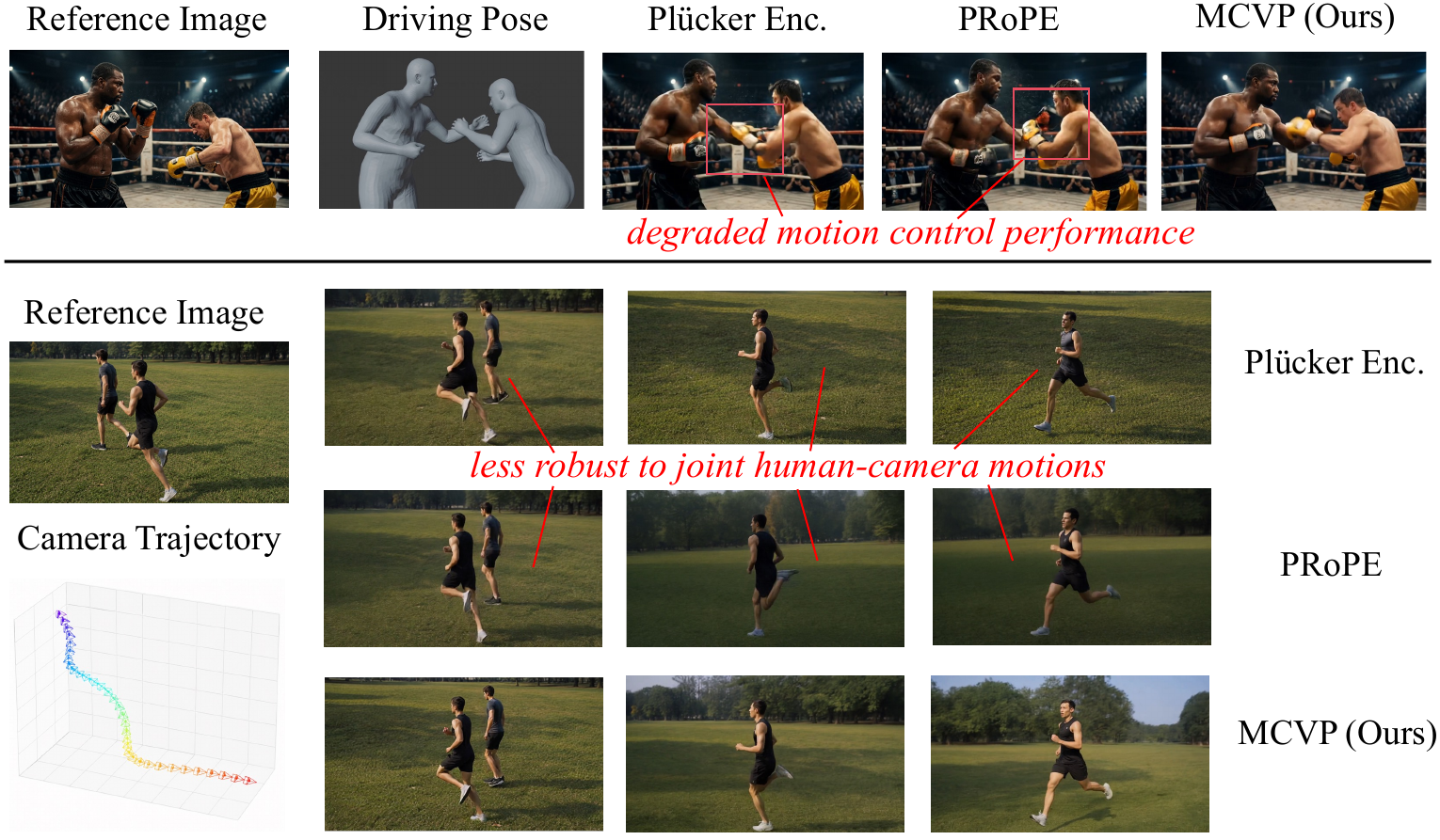}
  \caption{Heterogeneous motion-camera controls degrade video generation quality.}
  \label{fig:ablation}
  \vspace{-1.5em}
\end{figure}
\vspace{-0.5em}
\subsection{Ablation Study}

% Top: for the same reference image and driving pose, Pl\"ucker encoding and PRoPE degrade the articulated interaction, whereas MCVP preserves coherent body motion and contact. Bottom: when human and camera motions occur jointly, the parametric camera encodings are less robust to coupled changes in subject position and viewpoint, while MCVP maintains coherent motion and more faithfully follows the target camera trajectory.

\noindent\textbf{Unified Visual Conditioning.}
Table~\ref{tab:ablation} compares different motion-camera conditioning designs. We use Wan2.2 conditioned with pure motion sequence as baseline. Pl\"ucker encoding \cite{he2024cameractrl} and PRoPE~\cite{li2026cameras} make camera control possible, but they combine visual human-motion cues with heterogeneous parametric camera signals, which weakens the model's ability to separate body motion from viewpoint change. MCVP performs best across the ablation metrics and also improves motion and image-quality measurements over the no-camera baseline. Figure~\ref{fig:ablation} further shows that parametric encodings degrade contact and struggle under joint subject-camera motion, whereas MCVP preserves coherent articulation and viewpoint evolution.

\noindent\textbf{Training Components.}
Table~\ref{tab:ablation} also validates Shifted RoPE and MCVP dropout. Shifted RoPE keeps proxy tokens distinguishable from video tokens, while dropout improves robustness to imperfect conditioning. Their combination yields the best overall behavior.
\vspace{-0.5em}
\section{Conclusion}
We propose \textbf{UniMoCa}, a simple and effective framework that uses \textbf{MCVP} to represent human motion and camera trajectories in a unified visual space for joint motion and camera control in human video generation. Experiments show strong motion and camera control results, outperforming open-source baselines and remaining competitive with leading commercial systems.


\bibliography{aaai2027}

\end{document}